\documentclass[11pt,a4paper]{article}
\usepackage{amsmath}
\usepackage{amsfonts}

\usepackage[T1]{fontenc}
\usepackage[letterpaper]{geometry}
\usepackage[acceptedWithA]{tacl2018v2}
\usepackage[mathscr]{euscript}
\usepackage{multirow}
\usepackage{url}

\usepackage{times}
\usepackage{latexsym}
\usepackage{booktabs}
\usepackage{todonotes}
\usepackage{enumitem}
\usepackage{forest} % for tree DRSs
\usepackage{listings} % for amr
\usepackage{calc}	  % calculations over tikz coordinates	
\usetikzlibrary{calc} % calculations over tikz coordinates,
\usetikzlibrary{arrows.meta} % for nice arrows
\usepackage{tikz}     % drawing nodes
\usetikzlibrary{arrows} % tikz arrows
\usepackage{pifont} % arrow in amr to clausal form translation
\usepackage{booktabs,multirow} % nice tables
\usepackage{array,ragged2e}    % newcolumntype
\usepackage[normalem]{ulem} % underline and stikeout
\usepackage{subcaption}
\usepackage{relsize} % relative size of font
\usepackage{mymacros} % user's macros
\usepackage{multicol} % multicolumn structure

\usepackage{pdfcolfoot} % prevents footnotes to mess colouring
\newcommand{\mytilde}{\raise.17ex\hbox{$\scriptstyle\mathtt{\sim}$}}

\newcommand{\ttboxlab}[1]{\texttt{#1}}
\newcommand{\drgvar}[1]{\textbf{#1}}

\newcommand{\wsp}{\hspace{3mm}}
\newcommand{\posn}[2]{#1\kern-0.15em.\kern-0.15em#2}
\newcommand{\bvar}[1]{$[$#1$]$}
\newcommand{\evar}[1]{$\langle$#1$\rangle$}

\title{Exploring Neural Methods for Parsing\\ Discourse Representation Structures}
\author{
 Rik van Noord \qquad Lasha Abzianidze \qquad Antonio Toral \qquad Johan Bos \\
 Center for Language and Cognition, University of Groningen \\
  {\sf \{r.i.k.van.noord, l.abzianidze, a.toral.ruiz, johan.bos\}@rug.nl} \\
}

\date{}

\begin{document}
\maketitle
\begin{abstract}
Neural methods have had several recent successes in semantic parsing, though they have yet to face the challenge of producing meaning representations based on formal semantics.
We present a sequence-to-sequence neural semantic parser that is able to produce Discourse Representation Structures (DRSs) for English sentences with high accuracy, outperforming traditional DRS parsers. To facilitate the learning of the output, we represent DRSs as a sequence of flat clauses and introduce a method to verify that produced DRSs are well-formed and interpretable.
We compare models using characters and words as input and see (somewhat surprisingly) that the former performs better than the latter. We show that eliminating variable names from the output using De Bruijn-indices
increases parser performance. 
Adding silver training data boosts performance even further.
\end{abstract}

\section{Introduction}

Semantic parsing is the task of mapping a natural language expression to an interpretable meaning representation. Semantic parsing used to be the domain of symbolic and statistical approaches \cite{pereirashieber:1987,zelle1996learning,BlackburnBos:2005}. Recently however, neural methods, and in particular sequence-to-sequence models, have been successfully applied to a wide range of semantic parsing tasks. These include code generation \cite{ling:16}, question-answering  \cite{dong-lapata:16,he2016character} and Abstract Meaning Representation parsing \cite{konstas:17}.
Since these models have no intrinsic knowledge of the structure (tree, graph, set) they have to produce, recent work also focused on structured decoding methods, creating neural architectures that always output a graph or a tree \cite{buys:17,alvarez:17}. These methods often outperform the more general sequence-to-sequence models but are tailored to specific meaning representations.

\begin{figure}[!t]
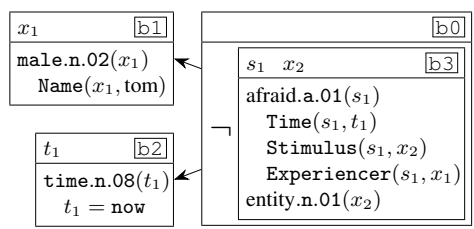

\centering
\parbox{0pt}{
\begin{tabbing}
12\=5\=\kill
{Raw input}:\\
\>\>\textsmaller[1]{\texttt{Tom isn't afraid of anything.}}
\\[0mm]
{System output of a DRS in a clausal form}:\\%
\>\>\textsmaller[2]{\texttt{%
    \begin{tabular}[t]{@{}l@{\kern5mm}l}
    \drgvar{b1} REF \drgvar{x1}         
        & \drgvar{b3} REF \drgvar{s1}\\
    \drgvar{b1} male "\posn{n}{02}" \drgvar{x1}
        & \drgvar{b3} Time \drgvar{s1} \drgvar{t1}\\
    \drgvar{b1} Name \drgvar{x1} "tom"
        & \drgvar{b3} Experiencer \drgvar{s1} \drgvar{x1}\\
    \drgvar{b2} REF \drgvar{t1}
        & \drgvar{b3} afraid "\posn{a}{01}" \drgvar{s1}\\
    \drgvar{b2} EQU \drgvar{t1} "now"
        & \drgvar{b3} Stimulus \drgvar{s1} \drgvar{x2}\\
    \drgvar{b2} time "\posn{n}{08}" \drgvar{t1}
        & \drgvar{b3} REF \drgvar{x2}\\
    \drgvar{b0} NOT \drgvar{b3}
        & \drgvar{b3} entity "\posn{n}{01}" \drgvar{x2}
    \end{tabular}}}    
\\[1mm]
{The same DRS in a box format}:\\[1mm]
\>\>\scalebox{.9}{%
    \begin{drstree}{-1}{4mm}{4mm}
    [{\pdrs{b0}{}{\textlarger[3]{$\neg$}
          \pdrs[c]{b3}{$s_1$ ~ $x_2$}{
            afraid$\sym{.a.01}(s_1)$\\
            \wsp$\sym{Time}(s_1, t_1)$\\ 
            \wsp$\sym{Stimulus}(s_1, x_2)$\\
            \wsp$\sym{Experiencer}(s_1, x_1)$\\
            entity$\sym{.n.01}(x_2)$}}}
        [{\pdrs{b1}{$x_1$}{
            $\sym{male.n.02}(x_1)$\\
            \wsp$\sym{Name}(x_1, \text{tom})$}}
        ]
        [{\pdrs{b2}{$t_1$}{
            $\sym{time.n.08}(t_1)$\\
            \wsp$t_1 = \sym{now}$}} 
        ]
    ]
    \end{drstree}
    }
\end{tabbing}
}
\caption{DRS parsing in a nutshell: given a raw text, a system has to generate a DRS in the clause format, a flat version of the standard box notation. The semantic representation formats are made more readable by using various letters for variables:
the letters \texttt{x}, \texttt{e}, \texttt{s}, and \texttt{t} are used for discourse referents denoting individuals, events, states and time, respectively, while \texttt{b} is used for variables denoting DRS boxes.
}
\label{fig:semantic_parsing}
\end{figure}

This paper will focus on parsing Discourse Representation Structures (DRSs) proposed in Discourse Representation Theory (DRT), a well-studied formalism developed in formal semantics \cite{kamp:drt,van_der_sandt:92,kampreyle:drt,asher:drt,muskens:cdrt,eijckkamp:drt,kadmon:drt,asherlascarides}, dealing with many semantic phenomena: quantifiers, negation, scope ambiguities, pronouns, presuppositions, and discourse structure (see Figure~\ref{fig:semantic_parsing}). 
DRSs are recursive structures and form therefore a challenge for sequence-to-sequence models because they need to generate a well-formed structure and not something that looks like one but is not interpretable. 

The problem that we try to tackle bears similarities with the recently introduced task of mapping sentences to an Abstract Meaning Representation (AMR, \citealt{amr:13}). But there are notable differences between DRS and AMR.  
Firstly, DRSs contain scope, which results in a more linguistically motivated treatment of modals, quantification, and negation.
And secondly, DRSs contain a substantially higher number of variable bindings (reentrant nodes in AMR terminology), which are challenging for learning \cite{damonte:17}.

DRS parsing has been attempted already in the 1980s for small fragments of English \cite{johnsonklein,wadaasher}. Wide-coverage DRS parsers based on supervised machine learning emerged later \cite{step2008:boxer,le:12,boxer,neural_drs_gmb:18}. 
The objectives of this paper are to apply neural methods to DRS parsing. In particular, we are interested in answers to the following questions:

\begin{enumerate}[itemsep=-1.5mm, topsep=2mm]
    \item Are sequence-to-sequence models able to produce formal meaning representations (DRSs)?
    \item What is better for input: sequences of characters or sequences of words; does tokenization help; and what kind of casing is best used?
    \item What is the best way of dealing with variables that occur in DRSs?
    \item Does adding silver data increase the performance of the neural parser?
    \item What parts of semantics are learned and what parts of semantics are still challenging?
\end{enumerate}

\noindent
We make the following contributions to semantic parsing:\footnote{The code is available here: \url{https://github.com/RikVN/Neural_DRS}.} 
(a) The output of our parser consists of interpretable scoped meaning representations, guaranteed by a specially designed checking tool (Section\,\ref{sec:method});
(b) We compare different methods of representing input and output in Section~\ref{sec:rep_exp};
(c) We show in Section~\ref{sec:exp_data} that employing additional, non-gold standard data can improve performance;
(d) We perform a thorough analysis of the produced output and compare our methods to symbolic/statistical approaches (Section~\ref{sec:discussion}).  

%███████ Semantic Parsing ███████
%███████

%████████████ DRS ████████████ 
\section{Discourse Representation Structures}
\label{sec:drs}

\subsection{The Structure of DRS}

DRSs are meaning representations introduced by DRT \cite{kampreyle:drt}.
In general, a DRS can be seen as an ordered pair $\langle \mathcal{A}, l\!:\!B \rangle$, where $\mathcal{A}$ is a set of presuppositional DRSs, and $B$ a DRS with a label $l$. The presuppositional DRSs $\mathcal{A}$ can be viewed as propositions that need to be anchored in the context in order to make the main DRS $B$ true, where presuppositions comprise anaphoric phenomena too \cite{van_der_sandt:92,geurts:pp,beaver:ppidrt}.

DRSs are either elementary DRSs or segmented DRSs. 
An elementary DRS is an ordered pair of a set of discourse referents and a set of conditions. 
There are basic conditions and complex conditions. A basic condition is a predicate applied to constants or discourse referents while a complex condition can introduce boolean operators ranging over DRSs (negation, conditionals, disjunction).  
Segmented DRSs capture discourse structure by connecting two units of discourse by a discourse relation \cite{asherlascarides}.

%██████████
\subsection{Annotated Corpora}
\label{ssec:annot_corpora}

Despite a long tradition of formal interest in DRT, it is only since recently that textual corpora annotated with DRSs have been made available. The Groningen Meaning Bank (GMB) is a large corpus with DRS annotation for mostly short English newspaper texts \cite{gmb:lrec,GMB:2017}. The DRSs in this corpus are produced by an existing semantic parser and then partially corrected. The DRSs in the GMB are therefore not gold standard.

A similar corpus is the Parallel Meaning Bank (PMB), that provides DRSs for English, German, Dutch and Italian sentences based on a parallel corpus \cite{PMBshort:2017}. 
The PMB, too, is constructed using an existing semantic parser, but a part of it is completely manually checked and corrected (i.e., gold standard). 
In contrast to the GMB, the PMB involves two major additions:
(a) its semantics are refined by modelling tense and employing semantic tagging \cite{Bjervaetal:16,semantic-tagset:17}, and (b) the non-logical symbols of the DRSs corresponding to concepts and semantic roles are grounded in WordNet \cite{wordnet} and VerbNet \cite{Bonial:11} respectively.

These above-mentioned additions make the DRSs of the PMB more fine-grained meaning representations.
For this reason we choose the PMB (over the GMB) as our corpus for evaluating our semantic parser. Even though the sentences in the current release of the PMB are relatively short, they contain many hard semantic phenomena that a semantic parser has to deal with: pronoun resolution, quantifiers, scope of modals and negation, multi-word expressions, word senses, semantic roles, presupposition, tense, and discourse relations.
As far as we know, we are the first that employs the PMB corpus for semantic parsing.

%██████████
\subsection{Formatting DRSs with Boxes and Clauses}

\begin{figure}[t!]
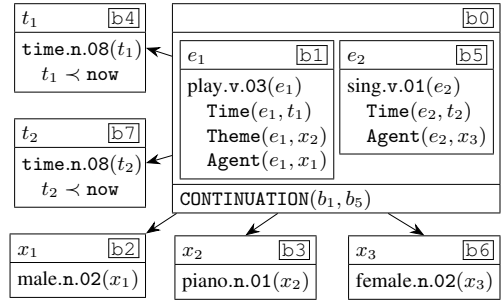

\hspace{-4mm}
\scalebox{0.95}{\begin{tabular}{c}
\mbox{\href{http://pmb.let.rug.nl/explorer/explore.php?part=00&doc_id=3008&type=der.xml}{\small 00/3008}}: 
\textsmaller[1]{\texttt{He played the piano and she sang.}}
\\[-2mm]
\textsmaller[2]{\texttt{
\begin{tabular}[t]{l@{\kern5mm}l}\toprule
\drgvar{b0} DRS \drgvar{b1} &
\drgvar{b0} DRS \drgvar{b5}\\
\drgvar{b2} REF \drgvar{x1} &
\drgvar{b6} REF \drgvar{x3}\\
\drgvar{b2} male "\posn{n}{02}" \drgvar{x1} &
\drgvar{b6} female "\posn{n}{02}" \drgvar{x3}\\
\drgvar{b1} REF \drgvar{e1} & 
\drgvar{b5} REF \drgvar{e2}\\
\drgvar{b1} play "\posn{v}{03}" \drgvar{e1} &
\drgvar{b5} sing "\posn{v}{01}" \drgvar{e2}\\
\drgvar{b1} Agent \drgvar{e1} \drgvar{x1} &
\drgvar{b5} Agent \drgvar{e2} \drgvar{x3}\\
\drgvar{b1} Theme \drgvar{e1} \drgvar{x2} &
\drgvar{b5} Time \drgvar{e2} \drgvar{t2}\\
\drgvar{b3} REF \drgvar{x2} &
\drgvar{b7} REF \drgvar{t2}\\
\drgvar{b3} piano "\posn{n}{01}" \drgvar{x2} & 
\drgvar{b7} TPR \drgvar{t2} "now"\\
\drgvar{b4} REF \drgvar{t1} & 
\drgvar{b7} time "\posn{n}{08}" \drgvar{t2}\\
\drgvar{b4} time "\posn{n}{08}" \drgvar{t1} &
\drgvar{b0} CONTINUATION \drgvar{b1} \drgvar{b5}\\
\drgvar{b4} TPR \drgvar{t1} "now" &
\drgvar{b1} Time \drgvar{e1} \drgvar{t1}\\
\bottomrule
\end{tabular}
}}
\\\noalv{2mm}
\scalebox{.9}{
\begin{drstree}{-1}{4mm}{3mm}
[,phantom, s sep=3.2mm
[{\psdrs{b0}{}{
    \pdrs{b1}{$e_1$}{
        play$\sym{.v.03}(e_1)$\\
        \wsp$\sym{Time}(e_1, t_1)$\\
        \wsp$\sym{Theme}(e_1, x_2)$\\
        \wsp$\sym{Agent}(e_1, x_1)$}
    \pdrs{b5}{$e_2$}{
        sing$\sym{.v.01}(e_2)$\\
        \wsp$\sym{Time}(e_2, t_2)$\\
        \wsp$\sym{Agent}(e_2, x_3)$}}
    {$\sym{CONTINUATION}(b_1,b_5)$}}, name=b0, s sep=4mm
    [{\pdrs{b4}{$t_1$}{
        $\sym{time.n.08}(t_1)$\\
        \wsp$t_1 \prec \sym{now}$}}
    ]
    [{\pdrs{b7}{$t_2$}{
        $\sym{time.n.08}(t_2)$\\
        \wsp$t_2 \prec \sym{now}$}}
    ]    
    [{\pdrs{b2}{$x_1$}{
        male$\sym{.n.02}(x_1)$}}, child anchor = north east
    ]
]    
    [{\pdrs{b3}{$x_2$}{
        piano$\sym{.n.01}(x_2)$}}, name=b3, xshift=-28.3mm
    ]
    [{\pdrs{b6}{$x_3$}{
        female$\sym{.n.02}(x_3)$}}, name=b6
    ]    
]
\draw[-{Stealth[scale=1.3]}] ($(b0.south)-(.9,0)$) -- (b3.north);
\draw[-{Stealth[scale=1.3]}] ($(b0.south)+(.8,0)$) -- (b6.north);
\end{drstree}}
\end{tabular}
}
\caption{A segmented DRS. Discourse relations are formatted with uppercase characters.}
\label{fig:drs_00_3008}
\end{figure}

The usual way to represent DRSs is the well-known box-format.
In order to facilitate reading a DRS with unresolved presuppositions, it can be depicted as a network of boxes, where a non-presuppositional (i.e., main) DRS $l\!:\!B$ is connected to the presuppositional DRSs $\mathcal{A}$ with arrows.
Each box comes with a unique label and has two rows.
In case of elementary DRSs these rows contain discourse referents in the top row and conditions in the bottom row (Figure~\ref{fig:semantic_parsing}). 
A segmented DRS has a row with labelled DRSs and a row with discourse relations (Figure~\ref{fig:drs_00_3008}).  

%███████ DRS ███████
%███████

The DRS in Figure~\ref{fig:semantic_parsing} consists of a main box \ttboxlab{b0} and two presuppositional boxes, \ttboxlab{b1} and \ttboxlab{b2}. Note that \ttboxlab{b0} has no discourse referents but introduces negation via a single condition $\neg$\ttboxlab{b3} with a nested box \ttboxlab{b3}.
The conditions of \ttboxlab{b3} represent unary and binary relations over discourse referents that are introduced either by \ttboxlab{b3} or the presuppositional DRSs.

A clausal form is another way of formatting DRSs. 
It represents a DRS as a set of clauses (see Figure~\ref{fig:semantic_parsing} and \ref{fig:drs_00_3008}).
This format is better suitable for machine learning than the box-format as it has a simple, flat structure and facilitates partial matching of DRSs which is useful for evaluation \cite{pmb-LREC:18}.
Conversion from the box-notation to the clausal form and vice versa is transparent: discourse referents, conditions, and discourse relations in the clausal form are preceded by the label of the box they occur in.
Notice that the variable letters in the semantic representations are automatically set and they simply serve for readability purposes. 
Throughout the experiments described in this paper, we employ clausal form DRSs.  

%██████████ Method ██████████ 
\section{Method}
\label{sec:method}

%██████████
\subsection{Annotated Data}
\label{ssec:data}

We use the English DRSs from release 2.1.0 of the PMB \cite{PMBshort:2017}.%
\footnote{\url{http://pmb.let.rug.nl/data.php}} 
The release suggests to use the parts 00, 10, 20 and 30 as the development set, resulting in 3,998 train and 557 development instances. 
Basic statistics are shown in Table~\ref{pmb:stats}, while the number of occurrences of some of the semantic phenomena mentioned in Section~\ref{ssec:annot_corpora} are given in Table~\ref{pmb:semstats}. 

\begin{table}[t]
\centering
\scalebox{0.90}{
\begin{tabular}{lrrc}
\toprule
                     & \textbf{Sentences} & \textbf{Tokens} & \textbf{Avg tok/sent} \\ \midrule
\textbf{Gold train}            & 3,998             & 24,917     & 6.2       \\
\textbf{Gold test}              & 557              & 3,180  &  5.7          \\
\textbf{Silver}              & 73,778           & 638,610   & 8.7     \\ \bottomrule  
\end{tabular}
}
\caption{Number of documents, sentences and tokens for the English part of PMB release 2.1.0. Note that the number of tokens is based on the PMB tokenization, treating multi-word expressions as a single token.}
\label{pmb:stats}
\end{table}

\begin{table}[t]
\centering 
\scalebox{.92}{
\begin{tabular}{lrrr}
\toprule
\textbf{Phenomenon} & \textbf{Train} & \textbf{Test} & \textbf{Silver} \\  \midrule
\textbf{negation \& modals} 
    & 442 & 73 & 17,527\\
\textbf{scope ambiguity}    
    & $\approx$67 & 15 & $\approx$3,108\\
\textbf{pronoun resolution} 
    & $\approx$291 & 31 & $\approx$3,893\\
\textbf{discourse rel. \& imp.} 
    & 254 & 33 & 16,654\\
\textbf{embedded clauses} 
    & $\approx$160  & 30    & $\approx$46,458\\

\bottomrule 
\end{tabular}
}
\caption[]{Counts of relevant semantic phenomena for PMB release 2.1.0.\footnotemark{}
These phenomena are described and further discussed in Section\,\ref{sec:manual}.}
\label{pmb:semstats}
\end{table}

Since this is a rather small training set, we tune our model using 10-fold cross-validation (CV) on the training set, instead of tuning on a separate development set. This means that we will use the suggested development set as a test set (and refer to it as such). When testing on this set, we train a model on all available training data.%
The employed PMB release also comes with ``silver'' data, namely, 71,308 DRSs that are only partially manually corrected. 
In addition, we employ the DRSs from the silver data but without the manual corrections, which makes them ``bronze'' DRSs following the PMB terminology. Our experiments will initially use only the gold standard data, after which we will employ the silver or bronze data to further push the score of our best systems.%
\footnotetext{The phenomena are automatically counted based on clausal forms. The counting algorithm does not guarantee the exact number for certain phenomena, though it returned the exact counts of all the phenomena on the test data except the pronoun resolution (30).}

%██████████
\subsection{Clausal Form Checker}
\label{ssec:clf_checker}

The clausal form of a DRS needs to satisfy a set of constraints in order to correspond to a semantically interpretable DRS, i.e., translatable into a first-order logic formula without free occurrences of a variable \cite{kampreyle:drt}.
For example, all discourse referents need to be explicitly introduced with a \texttt{REF} clause to avoid free occurrences of variables.

We implemented a clausal form checker that validates the clausal form if and only if it represents a semantically interpretable DRS.
Distinguishing box variables from entity variables is crucial for the validity checking, but automatically learned clausal forms are not expected to differentiate variable types.
First, the checker separately parses each clause in the form to induce variable types based on the fixed set of comparison and DRS operators.
After typing all the variables, the checker verifies whether the clauses collectively correspond to a DRS with well-formed semantics. 
For each box variable in a discourse relation, existence of the corresponding box inside the same segmented DRS is checked.
For each entity variable in a condition, an introduction of the binder (i.e., accessible) discourse variable is found.
The goal of these two steps is to prevent free occurrences of variables in DRSs.   
While binding the entity variables, necessary accessibility relations between the boxes are induced.
In the end, the checker verifies the transitive closure of the induced accessibility relation on loops and checks existence of a unique main box of the DRS.  

The checker is applied to every automatically obtained clausal form.
If a clausal form fails the test, it is considered as ill-formed and will not have a single clause matched with the gold standard when calculating the F-score.

%██████████
\subsection{Evaluation}
\label{ssec:evaluation}

A DRS parser is evaluated by comparing its output DRS to a gold standard DRS using the Counter tool \cite{pmb-LREC:18}. Counter calculates an F-score over matching clauses. Since variable names are meaningless, obtaining the matching clauses essentially is a search for the best variable mapping between two DRSs. Counter tries to find this mapping by performing a hill-climbing search with a predefined number of restarts to avoid getting stuck in a local optimum, which is similar to the evaluation system \textsc{smatch} \cite{smatch:13} for AMR parsing.%
\footnote{Counter ignores \texttt{REF} clauses in the calculation of the F-score since they are usually redundant and therefore inflate the final score \cite{pmb-LREC:18}.} 
Counter generalises over WordNet synsets, i.e., a system is not penalised for predicting a word sense that is in the same synset as the gold standard word sense. 

To calculate whether there is a significant difference between two systems, we perform approximate randomization \cite{random-appr:89} with $\alpha$ = $0.05$, \emph{R} = $1000$ and $F(model_{1}) > F(model_{2})$ as test statistic for each individual DRS pair. 

%██████████
\subsection{Neural Architecture}
\label{sec:parameters}

\begin{figure}[!t]
  \centering
  \includegraphics[scale=0.425]{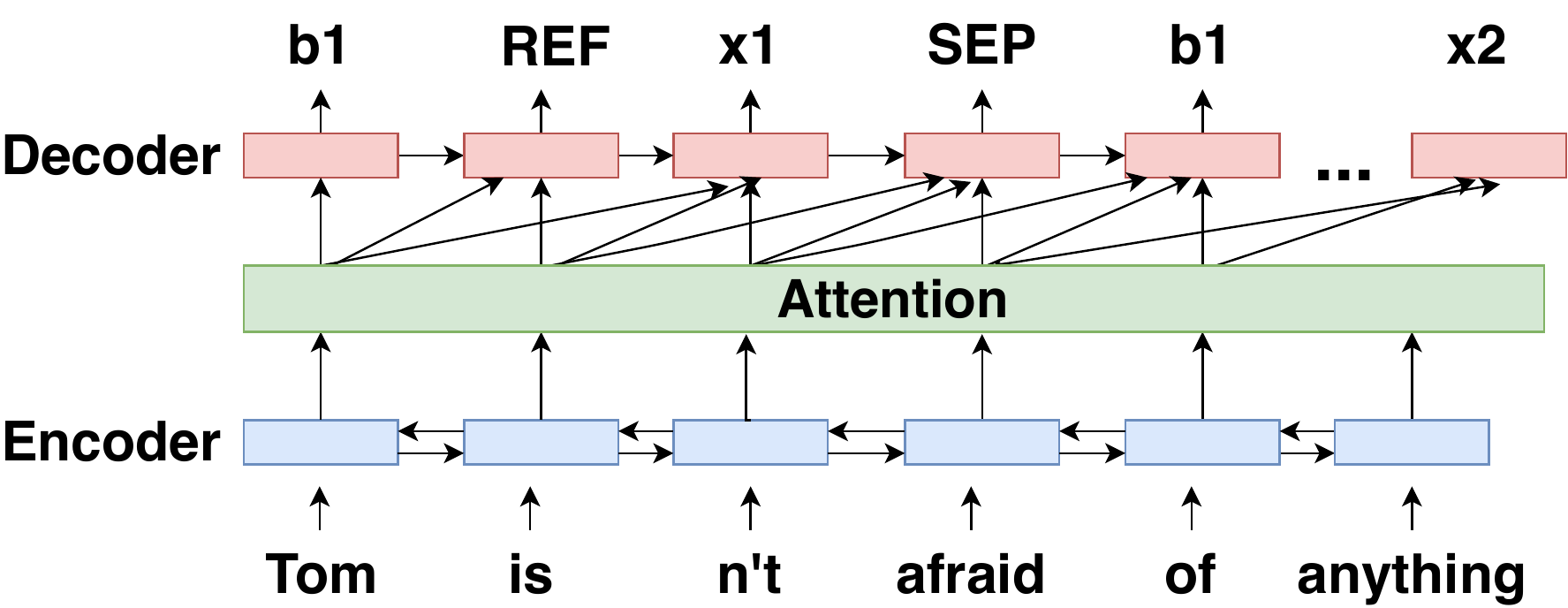}
  \caption{\label{fig:seq2seq}The sequence-to-sequence model with word-representation input. \texttt{SEP} is used as a special character to separate clauses in the output.}
\end{figure}

We employ a recurrent sequence-to-sequence neural network (henceforth seq2seq) with two bidirectional LSTM layers and 300 nodes, implemented in OpenNMT \cite{opennmt:17}.
The network encodes a sequence representation of the natural language utterance, while the decoder produces the sequences of the meaning representation. We apply dropout \cite{dropout:14} between both the recurrent encoding and decoding layers to prevent overfitting, and use general attention \cite{luong15} to selectively give more weight to certain parts of the input sentence. An overview of the general framework of the seq2seq model is shown in Figure~\ref{fig:seq2seq}.

During decoding we perform beam search with length normalization, which in neural machine translation (NMT) is crucial to obtaining good results \cite{britz:17}. We experimented with a wide range of parameter settings, of which the final settings can be found in Table~\ref{tab:param}.

We opted against trying to find the best parameter settings for each individual experiment (next to impossible in terms of computing time necessary as a single 10-fold CV experiment takes 12 hours on GPU), but selected parameter settings that showed good performance for both the initial character and word-level representations (see Section~\ref{sec:rep_exp} for details). The parameter search was performed using 10-fold CV on the training set. Training is stopped when there is no more improvement in perplexity on the validation set, which in our case occurred after 13--15 epochs.

A powerful, well-known technique in the field of NMT is to use an ensemble of models during decoding \cite{sutskever:14,sennrich-wmt:16}. The resulting model averages over the predictions of the individual models, which can balance out some of the errors. In our experiments, we apply this method when decoding on the test set, but not for our experiments of 10-fold CV (this would take too much computation time).

\begin{table}[t]
\centering
\scalebox{0.815}{
\begin{tabular}{lrlr}
\toprule
\textbf{Parameter}   & \textbf{Value} & \textbf{Parameter}   & \textbf{Value}                 \\ \midrule
RNN-type             & LSTM           &  dropout              & 0.2     \\
encoder-type         & brnn           &  dropout type         & naive    \\
optimizer            & sgd            &  bridge               & copy    \\
layers               & 2              &  learning rate        & 0.7         \\
nodes                & 300            &  learning rate decay  & 0.7       \\
min freq source      & 3              &  max grad norm        & 5          \\
min freq target      & 3              &  beam size            & 10       \\
vector size          & 300            &  length normalisation & 0.9       \\ \bottomrule       
\end{tabular}
}
\caption{Parameters explored during training and testing with their final values. All other parameters have default values.}
\label{tab:param}
\end{table}

%██████████ Exp on Representation ██████████ 
\section{Experiments with Data Representations}
\label{sec:rep_exp}

This section describes the experiments we conduct regarding the data representations of the input (English sentences) and output (a DRS) during training.

\subsection{Between Characters and Words}

We first try two (default) representations: character-level and word-level. Most semantic parsers use word-level representations for the input, but as a result are often dependent on pre-trained word embeddings or anonymization of the input
\footnote{This is done to keep the vocabulary small. An example is to change all proper names to \texttt{NAME} in both the sentence and meaning representation during training. When producing output, the original names are restored by switching \texttt{NAME} with a proper name found in the input sentence \cite{konstas:17}.}
to obtain good results. Character-level models avoid this issue but might be at a higher risk of producing ill-formed output.

\paragraph{Character-based model}

In the character-level model, the input (an English sentence) is represented as a sequence of individual characters. The output (a DRS in clause format) is linearized, with special characters indicating spaces and clause separators. The semantic roles  (e.g. \texttt{Agent}, \texttt{Theme}), DRS operators (e.g. \texttt{REF}, \texttt{NOT}, \texttt{POS}) and deictic constants (e.g. \texttt{"now"}, \texttt{"speaker"}, \texttt{"hearer"}) are not represented as character sequences, but treated as compound characters, meaning that \texttt{REF} is not treated as a sequence of \texttt{R}, \texttt{E} and \texttt{F}, but directly as \texttt{REF}.
All proper names, WordNet senses, time/date expressions, and numerals are represented as character sequences.

\paragraph{Word-based model}

In the word-level model, the input is represented as a sequence of words, using spaces as a separator (i.e., the original words are kept). The output is the same as for the character-based model, except that the character sequences are represented as words.
We use pre-trained GloVe embeddings \cite{glove:14}%
\footnote{The Common Crawl version trained on 840 billion tokens, vector size 300.} 
to initialise the encoder and decoder representations. In the DRS representation, there are semantic roles and DRS operators that might look like English words, but should not be interpreted as such (e.g. \texttt{Agent}, \texttt{NOT}). These entities are removed from the set of pre-trained embeddings, so that the model will learn them from scratch (starting from a random initialization). 

\paragraph{Hybrid representations: BPE}

We do not necessarily have to restrict ourselves to using only characters or words as input representation. In NMT, byte-pair encoding (BPE, \citealt{sennrich-haddow-birch:2016:P16-12}) is currently the \emph{de facto} standard \cite{bojar-EtAl:2017:WMT1}. 
This is a frequency-based method that automatically finds a representation that is in between character and word-level. 
It starts out with the character-level format and then does a predefined number of merges of frequently co-occurring characters. Tuning this number of merges determines if the resulting representation is closer to character or word-level. We explore a large range of merges (1k--100k), while applying a corresponding set of pre-trained BPE embeddings \cite{bpe_pretrain:18}. However, none of the BPE experiments improved on the character-level or word-level score (F-scores between 57 and 68), only coming close when using a small number of merges (which is very close to character-level anyway). Therefore this technique was disregarded for further experiments.			
			
\paragraph{Combined char and word}			
			
There is also a fourth possible representation of the input: concatenating the character and word-level representations. This is uncommon in NMT due to the large size of the embedding space (hence their preference for BPE), but possible here since the PMB data contains relatively short sentences. We simply add the word embedding vector after the sequence of character-embeddings for each word in the input and still initialise these embeddings using the pre-trained GloVe embeddings. 

\paragraph{Representation results}

The results of the experiments (10-fold CV) for finding the best representation are shown in Table~\ref{tab:rep}. Character representations are clearly better than word representations, though the word-level representation produces fewer ill-formed DRSs. Both representations are maintained for our further experiments. 
Although the combination of characters and words did lead to a small increase in performance over characters only (Table~\ref{tab:rep}), this difference is not significant. Hence, this representation is discarded in further experiments described in this paper.

\begin{table}[!htb]
\centering
\scalebox{0.95}{
\begin{tabular}{lcccc}
\toprule
\textbf{Model} & \textbf{Prec} & \textbf{Rec} & \textbf{F-score} & \textbf{\%\,ill} \\ \midrule
Char           & 78.1 & 69.7 & 73.7  & 6.2            \\
Word           & 73.2 & 65.9 & 69.4   & 5.8             \\
Char + Word    & 78.9  & 69.7 & 74.0  & 7.5  \\ \bottomrule        
\end{tabular}
}
\caption{Evaluating different input representations. 
The percentage of ill-formed DRSs is denoted by  \%\,ill.}
\label{tab:rep}
\end{table}

%██████████ Variable renaming ██████████
\begin{figure*}[!t]
\hspace*{-3mm}
\centering
\begin{subfigure}{50mm}
\centering
\textsmaller[2]{\texttt{
\begin{tabular}[t]{@{}l@{}}
\midrule
\drgvar{b1} REF \drgvar{x1}\\
\drgvar{b1} male "\posn{n}{02}" \drgvar{x1}\\
\drgvar{b1} Name \drgvar{x1} "tom"\\
\drgvar{b2} REF \drgvar{t1}\\
\drgvar{b2} EQU \drgvar{t1} "now"\\
\drgvar{b2} time "\posn{n}{08}" \drgvar{t1}\\
\drgvar{b0} NOT \drgvar{b3}\\
\drgvar{b3} REF \drgvar{s1}\\
\drgvar{b3} Time \drgvar{s1} \drgvar{t1}\\
\drgvar{b3} Experiencer \drgvar{s1} \drgvar{x1}\\
\drgvar{b3} afraid "\posn{a}{01}" \drgvar{s1}\\
\drgvar{b3} Stimulus \drgvar{s1} \drgvar{x2}\\
\drgvar{b3} REF \drgvar{x2}\\
\drgvar{b3} entity "\posn{n}{01}" \drgvar{x2}\\
\midrule
\end{tabular}
}}
\caption{Standard naming}
\label{subfig:standard}
\end{subfigure}
\begin{subfigure}{50mm}
\centering
\textsmaller[2]{\texttt{
\begin{tabular}[t]{@{}l@{}}
\midrule
\drgvar{\$1} REF \drgvar{@1}\\
\drgvar{\$1} male "\posn{n}{02}" \drgvar{@1}\\
\drgvar{\$1} Name \drgvar{@1} "tom"\\
\drgvar{\$2} REF \drgvar{@2}\\
\drgvar{\$2} EQU \drgvar{@2} "now"\\
\drgvar{\$2} time "\posn{n}{08}" \drgvar{@2}\\
\drgvar{\$0} NOT \drgvar{\$3}\\
\drgvar{\$3} REF \drgvar{@3}\\
\drgvar{\$3} Time \drgvar{@3} \drgvar{@2}\\
\drgvar{\$3} Experiencer \drgvar{@3} \drgvar{@1}\\
\drgvar{\$3} afraid "\posn{a}{01}" \drgvar{@3}\\
\drgvar{\$3} Stimulus \drgvar{@3} \drgvar{@4}\\
\drgvar{\$3} REF \drgvar{@4}\\
\drgvar{\$3} entity "\posn{n}{01}" \drgvar{@4}\\
\midrule
\end{tabular}
}}
\caption{Absolute naming}
\label{subfig:absolute}
\end{subfigure}
\begin{subfigure}{50mm}
\centering
\textsmaller[2]{\texttt{
\begin{tabular}[t]{@{}l@{}}
\midrule
\drgvar{\bvar{NEW}} REF \drgvar{\evar{NEW}}\\
\drgvar{\bvar{0}} male "\posn{n}{02}" \drgvar{\evar{0}}\\
\drgvar{\bvar{0}} Name \drgvar{\evar{0}} "tom"\\
\drgvar{\bvar{NEW}} REF \drgvar{\evar{NEW}}\\
\drgvar{\bvar{0}} EQU \drgvar{\evar{0}} "now"\\
\drgvar{\bvar{0}} time "\posn{n}{08}" \drgvar{\evar{0}}\\
\drgvar{\bvar{NEW}} NOT \drgvar{\bvar{NEW}}\\
\drgvar{\bvar{0}} REF \drgvar{\evar{NEW}}\\
\drgvar{\bvar{0}} Time \drgvar{\evar{0}} \drgvar{\evar{-1}}\\
\drgvar{\bvar{0}} Experiencer \drgvar{\evar{0}} \drgvar{\evar{-2}}\\
\drgvar{\bvar{0}} afraid "\posn{a}{01}" \drgvar{\evar{0}}\\
\drgvar{\bvar{0}} Stimulus \drgvar{\evar{0}} \drgvar{\evar{1}}\\
\drgvar{\bvar{0}} REF \drgvar{\evar{NEW}}\\
\drgvar{\bvar{0}} entity "\posn{n}{01}" \drgvar{\evar{0}}\\
\midrule
\end{tabular}
}}
\caption{Relative naming}
\label{subfig:relative}
\end{subfigure}
\caption{Different methods of variable naming exemplified on the clausal form of Figure~\ref{fig:semantic_parsing}. For (c), positive numbers refer to introductions that have yet to occur, while negative numbers refer to known introductions. A zero refers to the previous introduction for that variable type.}
\label{fig:drs-examples}
\end{figure*}
%██████████

\subsection{Tokenization}

An interesting aspect of the PMB data is the way the input sentences are tokenized. In the data set, multi-word expressions are tokenized as single words, for example, ``New York'' is tokenized to ``New\mytilde{}York''. Unfortunately, most off-the-shelf tokenizers (e.g. the Moses tokenizer) are not equipped to deal with this. We experiment with using Elephant \cite{elephant}, a tokenizer that can be (re-)trained on individual data sets, using the tokenized sentences of the published silver and gold PMB data set.\footnote{Gold tokenization is available in the data set, but using this would not reflect practical applications of DRS parsing, as we want raw text as input for a realistic setting.} Simultaneously, we are interested in whether character-level models need tokenization at all, which would be a possible advantage of this type of representing the input text.

Results of the experiment are shown in Table~\ref{tab:res}. None of the two tokenization methods yielded a significant advantage for the character-level models, so they will not be employed further. The word-level models, however, did benefit from tokenization, but Elephant did not give us an advantage over the Moses tokenizer. Therefore, for word-level models, we will use Moses in our next experiments.

\subsection{Representing Variables}

So far we did not attempt to do anything special with the variables that occur in DRSs, as we simply tried to learn them as supplied in the PMB data set. Obviously, DRSs constitute a challenge for seq2seq models because of the high number of multiple occurrences of the same variables, in particular compared to AMR. AMR parsers do not deal well with this, since the reentrancy metric \cite{damonte:17} is among the lowest metrics for all AMR parsers that reported them or are publicly available \cite{clinAMR:17}. Moreover, for AMR, only 50\% of the representations contain at least one reentrant node, and only 20\% of the triples in AMR contain a reentrant node \cite{semdeep:17}, but for DRSs these are both virtually 100\%.
While seq2seq AMR parsers could get away with ignoring variables during training and reinstating them in a post-processing step, for DRSs this is unfeasible.

However, since variable names are chosen arbitrarily, they will be hard for a seq2seq model to learn. We will therefore experiment with two methods of rewriting the variables to a more general representation, distinguishing between box variables and discourse variables. Our first method (absolute) traverses down the list of clauses, rewriting each new variable to a unique representation, taking the order into account. The second method (relative) is more sophisticated; it rewrites variables based on when they were introduced, inspired by De Bruijn index \cite{debruijn:72}. We view box variables as introduced when they are first mentioned, while we take the \texttt{REF} clause of a discourse referent as their introduction. The two rewriting methods are illustrated in Figure~\ref{fig:drs-examples}. 

The results are shown in Table~\ref{tab:res}. For both characters and words, the relative rewriting method significantly outperforms the absolute method and the baseline, though the absolute method produces fewer ill-formed DRSs. 
Interestingly, the character-level model still obtains a higher F1-score compared to the word-level model, even though it produces more ill-formed DRSs.

\begin{table}[h]
\centering
\scalebox{0.8515}{
\setlength\tabcolsep{3.0pt}
\begin{tabular}{l|cc|cc}
                  & \multicolumn{2}{c|}{\textbf{Char parser}} & \multicolumn{2}{c}{\textbf{Word parser}} \\ \midrule
                  & \textbf{F1}   & \textbf{\%\,ill}   & \textbf{F1}   & \textbf{\%\,ill}   \\ \midrule
\textbf{Baseline (bs)} & 73.7         & 6.2                 &   69.4             & 5.8                 \\ \midrule
\textbf{Moses (mos)}    & 74.1         & 4.8              & 71.8 & 5.8                 \\
\textbf{Elephant (ele)} & 74.0             & 5.4                 & 71.1             &  7.5                \\ \midrule
\textbf{bs/mos + absolute (abs)} & 75.3          & 3.5               & 73.5          &  2.0             \\
\textbf{bs/mos + relative (rel)} & 76.3          & 4.2               & 74.2          &  3.1           \\  \midrule
\textbf{bs/mos + rel + lowercase} & 75.8 & 3.6 & 74.9 & 3.1           \\
\textbf{bs/mos + rel + truecase} & 76.2 & 4.0 & 73.3 & 3.3  \\
\textbf{bs/mos + rel + feature} & 76.9  & 3.7 & 74.9 & 2.9 
\end{tabular}
}
\caption{Results of the 10-fold CV experiments regarding tokenization, variable rewriting and casing. \emph{bs/mos} means that we use no tokenization for the character-level parser, while we use Moses for the word-level parser.}
\label{tab:res}
\end{table}

\subsection{Casing}

Casing is a writing device mostly used for punctuation purposes. On the one hand, it increases the set of characters (hence adding more redundant variation to the input). On the other hand, case can be a useful feature to recognise proper names as names of individuals are semantically analysed as presuppositions.
Explicitly encoding uppercase with a feature could therefore prevent us from including a named-entity recogniser, often used in other semantic parsers. Although we do not expect dealing with case is a major challenge, we try out different techniques to find an optimal balance between abstracting over input characters and parsing performance. The results, in Table~\ref{tab:res}, show that the feature works well for the character-level model, but for the word-level model, it does not outperform lowercasing. These settings are used in further experiments. 

\begin{figure}[!t]
  \centering
  \includegraphics[scale=0.420]{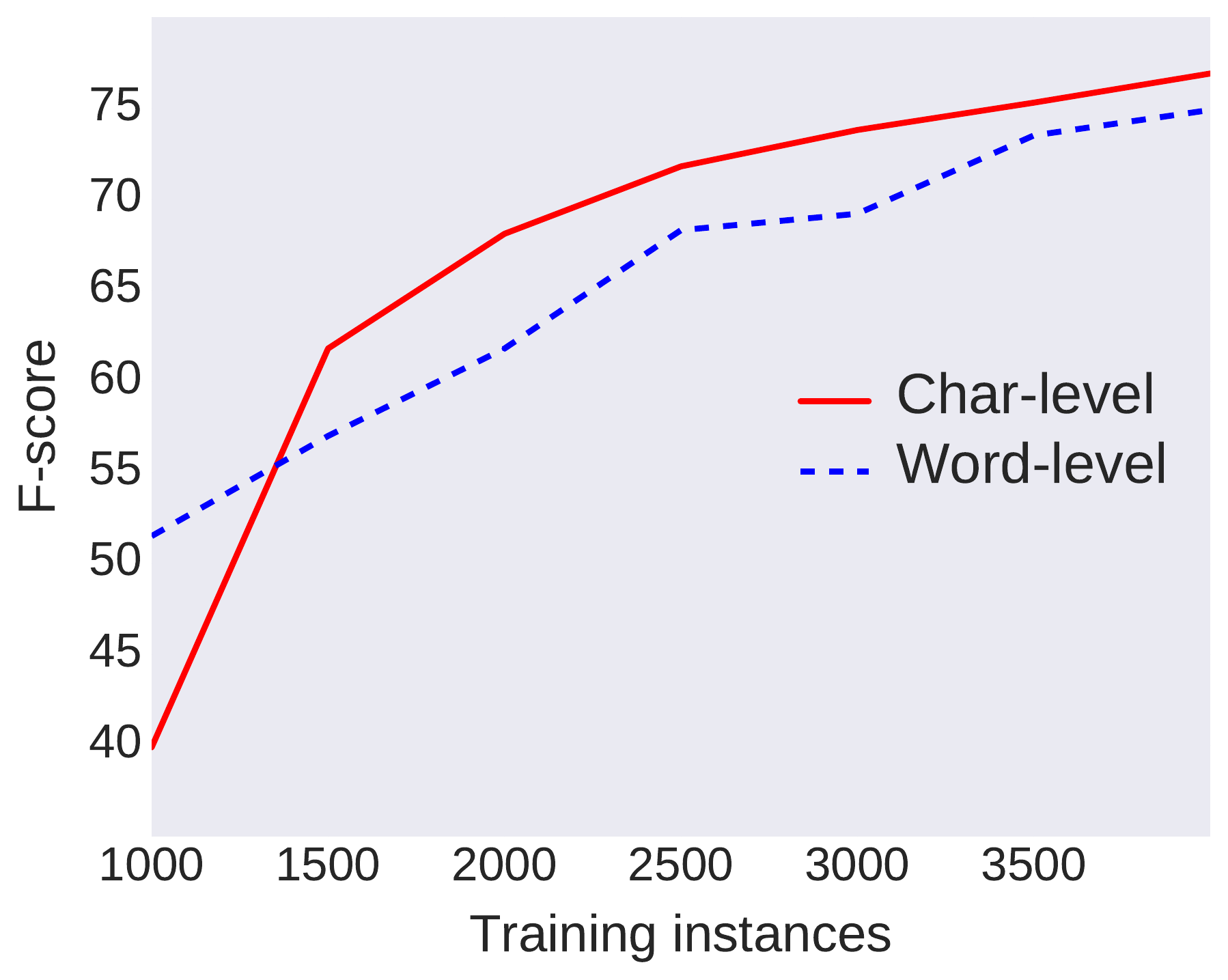}
  \caption{\label{fig:curve}Learning curve for different number of gold instances for both the character-level and word-level neural parsers (10-fold CV experiment for every 500 instances).}
\end{figure}

%██████████ Exp on additional data ██████████ 
\section{Experiments with Additional Data}
\label{sec:exp_data}

Since semantic annotation is a difficult and time-consuming task, gold standard data sets are usually relatively small. This means that semantic parsers (and data-hungry neural methods in particular) can often benefit from more training data. Some examples in semantic parsing are data recombination \cite{data-recomb:16}, paraphrasing \cite{berant2014semantic} or exploiting machine-generated output \cite{konstas:17}. However, before we do any experiments using extra training data, we want to be sure that we can still benefit from more gold training data. For both the character-level and word-level we plot the learning curve, adding 500 training instances at a time, in Figure~\ref{fig:curve}. For both models the F-score clearly still improves when using more training instances, which shows that there is at least the potential for additional data to improve the score.

For DRSs, the PMB-2.1.0 release already contains a large set of silver standard data (71,308 instances), containing DRSs that are only partially manually corrected. 
We then train a model on both the gold and silver standard data, making no distinction between them during training. After training we take the last model and restart the training on only the gold data, in a similar process as described in \newcite{konstas:17} and \newcite{clinAMR:17}. 
In general, restarting the training to fine-tune the weights of the model is a common technique in NMT \cite{denkowski:17}.

We are aware that there are many methods to obtain and employ additional data. However, our main aim is not to find the optimal method for DRS parsing, but to demonstrate that using additional data is indeed beneficial for neural DRS parsing. Since we are not further fine-tuning our model, we will show results on the test set in this section.

\begin{table}[!t]
\centering
\scalebox{0.93}{
\begin{tabular}{l|cc|cc}
                                 & \multicolumn{2}{c|}{\textbf{Char parser}} & \multicolumn{2}{c}{\textbf{Word parser}} \\ \midrule
           \textbf{Data}         & \textbf{F1}   & \textbf{\%\,ill}   & \textbf{F1}   & \textbf{\%\,ill}   \\ \midrule
\textbf{Best gold-only}          &  75.9            &   2.9               & 72.8                & 2.0                 \\
\textbf{\quad + ensemble}          & 77.9             &  1.8                & 75.1               & 0.9                 \\
\textbf{Gold + silver}       & 82.9              & 1.8                 &   82.7             &   1.1              \\ 
\textbf{\quad + ensemble}       & 83.6              & 1.3                 & 83.1               & 0.7                  \\ 
\end{tabular}
}
\caption{F1-score and percentage of ill-formed DRSs on the test set, for the experiments with the PMB-released silver data. The scores without using an ensemble are an average of five runs of the model.}
\label{tab:silver}
\end{table}

Table~\ref{tab:silver} shows the results of adding the silver data. 
This results in a large increase in performance, for both the character and word-level models. 
We are still reliant on manually annotated data, however, since without the gold data (so training on only the silver data), we score even lower than our baseline model (68.4 and 68.1 for the char and word parser). Similarly, we are reliant on the fine-tuning procedure, as we also score below our baseline models without it (71.6 and 71.0 for the char and word parsers, respectively).

We believe there are two possible factors that could explain why the addition of silver data results in such a large improvement: (i) the fact that the data is silver instead of bronze or (ii) the fact that a different DRS parser (Boxer, see Section~6), is used to create the silver data instead of our own parser. 

We conduct an experiment to find out the impact on performance of silver vs bronze and Boxer vs our parser. The results are shown in Table~\ref{tab:silver_impact}. 
Note that these experiments are performed to analyse the impact of the silver data, not to further push the score, meaning \emph{Silver (Boxer-generated)} is our final model that will be compared to other approaches in Section~6.

\begin{table}[!t]
\centering
\setlength{\tabcolsep}{3pt}
\scalebox{0.823}{
\begin{tabular}{l|cc|cc}
                                 & \multicolumn{2}{c|}{\textbf{Char parser}} & \multicolumn{2}{c}{\textbf{Word parser}} \\ \midrule
      \textbf{Data}                           & \textbf{F1}   & \textbf{\%\,ill}   & \textbf{F1}   & \textbf{\%\,ill}   \\ \midrule
\textbf{Silver (Boxer-generated)}             & 83.6              & 1.3                 & 83.1              & 0.7                 \\ 
\textbf{Bronze (Boxer-generated)}             & 83.8              & 1.1                 & 82.4              & 0.9\\
\textbf{Bronze (NN-generated)}             & 77.9              & 2.7                 & 74.5              & 2.2       \\
\textbf{\quad without ill-formed DRSs}  & 78.6              & 1.6                  & 74.9             & 0.9                   \\
\end{tabular}
}
\caption{Test set results of the experiments that analyse the impact of the silver data.}
\label{tab:silver_impact}
\end{table}

For (i), we compare the performance of the model trained on silver and bronze versions of the exact same documents (so leaving out the manual corrections). Interestingly, we score slightly higher for the character-level model with bronze than with silver (though the difference is not statistically significant), meaning that the extra manual corrections are not beneficial (in their current format). This suggests that the silver data is closer to bronze than to gold standard.

For (ii), we use our own best parser (without silver data) to parse the sentences in the PMB silver data release and use that as additional training data.\footnote{Note that we cannot apply the manual corrections, so in PMB terminology, this data is bronze instead of silver.} Since the silver data contains longer and more complicated sentences than the gold data, our best parser produces more ill-formed DRSs (13.7\% for char and 15.6\% for word). We can either discard those instances or still maintain them for the model to learn from. For Boxer this is not an issue since only 0.3\% of DRSs produced were ill-formed. We observe that a full self-training pipeline results in lower performance compared to using Boxer-produced DRSs. In fact, this does not seem to be beneficial over only using the gold standard data. Most likely, since Boxer combines symbolic and statistical methods, it learns very different things than our neural parsers, which in turn provides more valuable information to the model. A more detailed analysis on the difference in (semantic) output is performed in Section~\ref{sec:analysis} and \ref{sec:manual}. 
Removing ill-formed DRSs before training leads to higher F-scores for both the char and word parser, as well as a lower number of ill-formed DRSs.

\begin{table}[!t]
\centering
\scalebox{.9}{
\begin{tabular}{lccc} \toprule
                     & \textbf{Prec} & \textbf{Rec} & \textbf{F-score} \\ \midrule
\textbf{\textsc{spar}}        & 48.0   & 33.9    & 39.7           \\
\textbf{\textsc{sim-spar}}     &  55.6   & 57.9   & 56.8 \\ \midrule
\textbf{\textsc{amr2drs}}     & 43.3    & 43.0   & 43.2 \\
\textbf{Boxer}       & 75.7    & 72.9    & 74.3           \\ \midrule
\textbf{Neural Char} & 79.7    & 76.2   & 77.9       \\
\textbf{Neural Word} & 77.1   & 73.3   & 75.1      \\ 
\textbf{Neural Char + silver} & 84.7   & 82.4   & 83.6      \\
\textbf{Neural Word + silver} & 84.0   & 82.3   & 83.1      \\ 
 \bottomrule
              
\end{tabular}
}
\caption{Test set results of our best neural models compared to two baseline models and two parsers.}
\label{tab:comparison}
\end{table}

%██████████ Discussion ██████████ 
\section{Discussion}
\label{sec:discussion}

\subsection{Comparison}

In this section, we compare our best neural models (with and without silver data, see Table~\ref{tab:silver}) to two baseline systems and to two DRS parsers: \textsc{amr2drs} and Boxer.
\textsc{amr2drs} is a parser that obtains DRSs from AMRs by applying a set of rules \cite{pmb-LREC:18}, in our case using AMRs produced by the AMR parser of \newcite{clinAMR:17}.
Boxer is an existing DRS parser using a statistical CCG parser for syntactic analysis and a compositional semantics based on $\lambda$-calculus, followed by pronoun and presupposition resolution \cite{candcboxer:2007,step2008:boxer}.
\textsc{spar} is a baseline parser that outputs the same (fixed) default DRS for each input sentence. 
We implemented a second baseline model, \textsc{sim-spar}, which outputs, for each sentence in the test set, the DRS of the most similar sentence in the training set. This similarity is calculated by taking the cosine similarity of the average word embedding vector (with removed stopwords) based on the Glove embeddings \cite{glove:14}. 

Table~\ref{tab:comparison} show the result of the comparison.
The neural models comfortably outperform the baselines.
%, even the more sophisticated baseline of \textsc{sim-spar}. 
We see that both our neural models outperform Boxer by a large margin when using the Boxer labelled silver data. However, even without this dependence, the neural models perform significantly better than Boxer.
It is worth noting that the character-level model significantly outperforms the word-level model, even though it cannot benefit from pre-trained word embeddings and from a tokenizer. 

Concurrently with our work, a neural DRS parser has been developed by \newcite{neural_drs_gmb:18}. They use a customised neural
seq2seq model, which produces the DRS in three stages. It first predicts the general (deep) structure of the DRSs, after which the conditions and referents are filled in. 
Unfortunately, they train and evaluate their parser on annotated data from the GMB rather than from the PMB (see Section~2). This, combined with the fact that their work is contemporaneous to the current paper, makes it difficult to compare the approaches. However, we see no apparent reason why their method should not work on the PMB data.

\subsection{Analysis}
\label{sec:analysis}

An intriguing question is what our
models actually learn, and what parts of meaning are still challenging for neural methods. We do this in two ways, by performing an automatic analysis and by doing a manual inspection on a variety of semantic phenomena. Table~\ref{tab:auto_eval} shows an overview of the different automatic evaluation metrics we implemented with corresponding scores of the three models.

\begin{table}[!t]
\centering
\scalebox{0.90}{
\begin{tabular}{lccc}
\toprule
                          & \textbf{Char} & \textbf{Word} & \textbf{Boxer} \\ \midrule
\textbf{All clauses}    & 83.6 & 83.1 & 74.3         \\ \midrule
\textbf{DRS Operators}      & 93.2 & 93.3 & 88.0          \\
\textbf{VerbNet roles}          & 84.1 & 82.5 & 71.4         \\
\textbf{WordNet synsets}   & 79.7  & 79.4  & 72.5      \\
\textbf{\quad nouns}   & 86.1  & 88.5  & 82.5      \\ 
\textbf{\quad verbs, adverbs, adj.} & 65.1  & 58.7  & 49.3      \\ \midrule
\textbf{Oracle sense numbers}   & 86.7  & 85.7 & 78.1       \\
\textbf{Oracle synsets} & 90.7  & 90.9 & 83.8        \\
\textbf{Oracle roles}    & 87.4  & 87.2 & 82.0
\\ \bottomrule 
\end{tabular}
}
\caption{F-scores of fine-grained evaluation on the test set of the three semantic parsers.}
\label{tab:auto_eval}
\end{table}

The character- and word-level systems perform comparably in all categories except for VerbNet roles, where the character-based parser shows a clear advantage (1.6\% absolute).
The score for WordNet synsets is similar, but the word-level model has more difficulty predicting synsets that are introduced by verbs than for nouns. It is clear that the neural models outperform Boxer consistently on each of these metrics (partly because Boxer picks the first sense by default). What also stands out is the impact of the word senses: with a perfect word sense disambiguation module (oracle senses) large improvements can be gained for all three parsers.

It is interesting to look at what errors the model makes in terms of producing ill-formed output. For both the neural parsers, only about 2\% of the ill-formed DRSs are ill-formed because of a syntactic error in an individual clause (e.g. \texttt{b1 Agent x1}, where a fourth argument is missing), while all the other errors are due to a violated semantic constraint (see Section~\ref{ssec:clf_checker}). In other words, the produced output is a syntactically well-formed DRS but is not interpretable. 

\begin{figure}[!t]
\hspace*{-2.2mm}
  \includegraphics[width=0.49\textwidth]{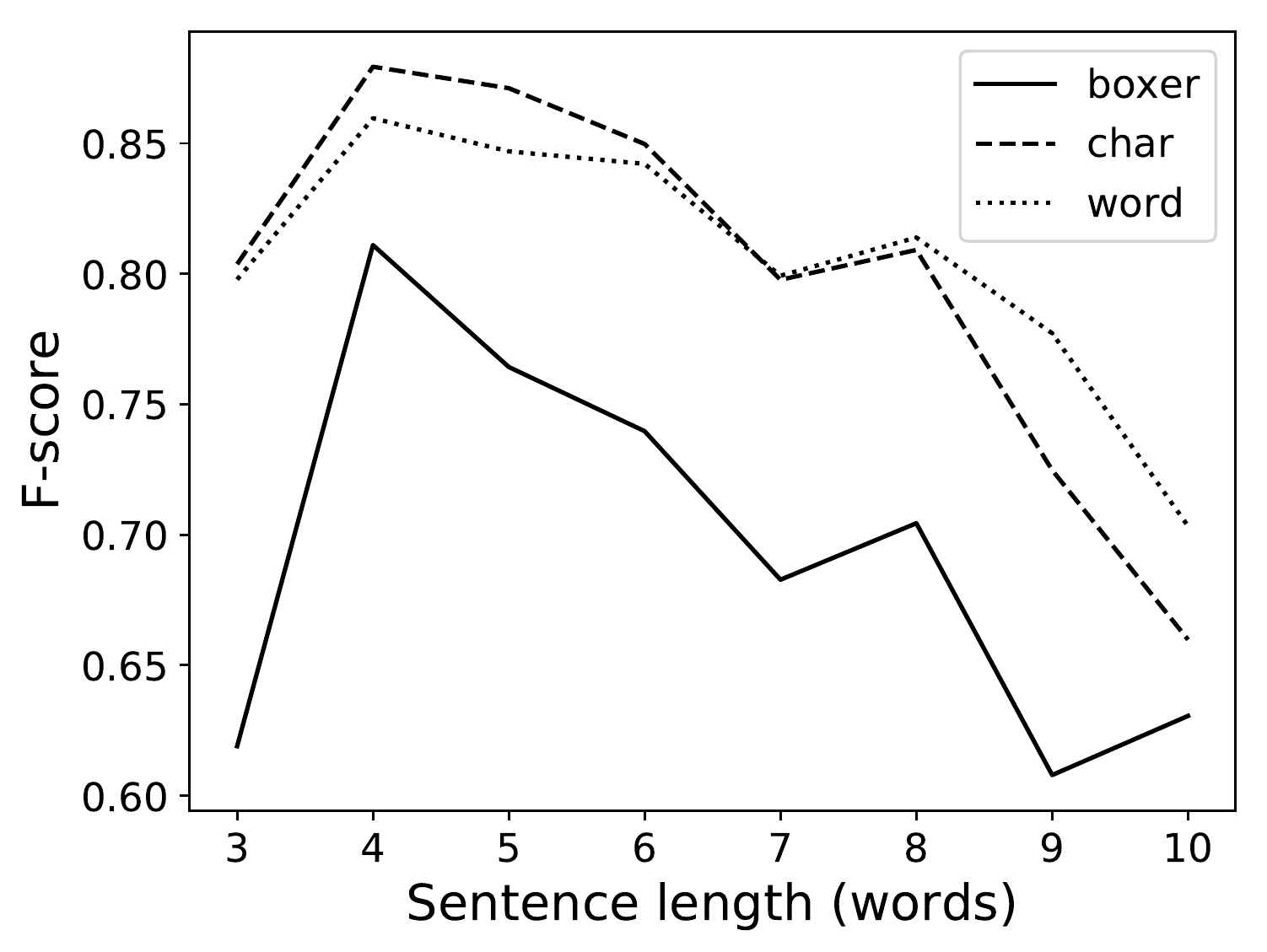}
  \caption{\label{fig:senlen}Performance of each parser for sentences of different length.}
\end{figure}

To find out how sentence length affects performance, we plot in Figure~\ref{fig:senlen} the mean F-score obtained by each parser on input sentences of different lengths, from 3 to 10 words.\footnote{Shorter and longer sentences are excluded as there are fewer than 10 input sentences for any such length, e.g. there are only 3 sentences that have 2 words.}
We observe that all the parsers degrade with sentence length.
To find out whether any of the parsers degrades significantly more than any other, 
we build a regression model, in which we predict the F-score 
using as predictors the parser (char, word and Boxer), the sentence length and the number of clauses produced. According to the regression model, (i) the performance of all the three systems decreases with sentence length, thus corroborating the trends shown in Figure~\ref{fig:senlen} and (ii) the interaction between parser and sentence length is not significant, i.e., none of the parsers decreases significantly more than any other with sentence length. 
The fact that the performance of the neural parsers degrades with sentence length is not surprising, since they are based on the seq2seq architecture, and models built on this architecture for other tasks, such as machine translation, have been shown to have the same issue~\cite{toral-sanchezcartagena:2017:EACLlong}.

\subsection{Manual Inspection}

\label{sec:manual}

The automatic evaluation metrics provide overall scores but do not capture how the models perform on certain semantic phenomena present in the DRSs. Therefore, we manually inspected the test set output of the three parsers for the semantic phenomena listed in Table~\ref{pmb:semstats}. 
Below we describe each phenomenon and explain how the parser output is evaluated on them.

The \textit{negation \& modals} phenomenon covers possibility (\texttt{POS}), necessity (\texttt{NEC}), and negation (\texttt{NOT}).
The phenomenon is considered successfully captured if an automatically produced clausal form has the clause with the modal operator and the main concept is correctly put under the scope of the modal operator. 
For example, to capture the negation in Figure\,\ref{fig:semantic_parsing}, the presence of {\small \texttt{\drgvar{b0}\;NOT\;\drgvar{b3}}} and {\small\texttt{\drgvar{b3}\;afraid\;"\posn{a}{01}"\;\drgvar{s1}}} is sufficient.   
\textit{Scope ambiguity} counts nested pairs of scopal operators such as possibility (\texttt{POS}), necessity (\texttt{NEC}), negation (\texttt{NOT}), and implication (\texttt{IMP}).
\textit{Pronoun resolution} checks if an anaphoric pronoun and its antecedent are represented by the same discourse referent.   
\textit{Discourse relation \& implication} involves determining a discourse relation or an implication with a main concept in each of their scopes (i.e., boxes).
For instance, to get the discourse relation in Figure\,\ref{fig:drs_00_3008} correctly, a clausal form needs to include 
{\small\texttt{\drgvar{b0}\;CONTINUATION\;\drgvar{b1}\;\drgvar{b5}}}, {\small\texttt{\drgvar{b1}\;play\;"\posn{v}{03}"\;\drgvar{e1}}}, 
and {\small\texttt{\drgvar{b5}\;sing\;"\posn{v}{01}"\;\drgvar{e2}}}.
Finally, the \textit{embedded clauses} phenomenon verifies whether the main verb concept of an embedded clause is placed inside the propositional box (\texttt{PRP}).
This phenomenon also covers control verbs: it checks if a controlled argument of a subordinate verb is correctly identified as an argument of a control verb.

The results of the semantic evaluation of the parsers on the test set is given in Table\,\ref{tab:manual_eval}. 
The character-level parser performs better than the word-level parser on all the phenomena except 
one.
Even though both our neural parsers clearly outperformed Boxer in terms of F-score, they perform worse than Boxer on the selected semantic phenomena. Although the differences are not big, Boxer obtained the highest score for four out of five phenomena.
This suggests that just the F-score is perhaps not good enough as an evaluation metric, or that the final F-score should perhaps be weighted towards certain clauses. For example, it is arguably more important to capture a negation correctly than tense. 
Our current metric only gives a rough indication about the contents, but not about the inferential capabilities of the meaning representation.

\begin{table}[t]
\centering
\scalebox{.915}{
\begin{tabular}{@{~}l@{\kern4mm}r@{\kern4mm}c@{\kern3mm}c@{\kern3mm}c@{~}}
\toprule
\textbf{Phenomenon} & \textbf{\#} & \textbf{Char} & \textbf{Word} & \textbf{Boxer}
\\ \midrule
\textbf{negation \& modals} &73 &\bf 0.90 & 0.81  & 0.89    \\
\textbf{scope ambiguity}    &15 & 0.73   & 0.57   & \bf 0.80     \\
\textbf{pronoun resolution} &31 & 0.84     & 0.77  &\bf 0.90      \\
\textbf{discourse rel. \& imp.} &33 & 0.64  & 0.67    & \bf 0.82  \\
\textbf{embedded clauses}   &30 & 0.77       & 0.70    & \bf 0.87    \\
\bottomrule 
\end{tabular}
}
\caption{Manual evaluation of the output of the three semantic parsers on several semantic phenomena. Reported numbers are accuracies.}
\label{tab:manual_eval}
\end{table}

%██████████ Conclusion ██████████ 
\section{Conclusions and Future Work}
\label{sec:conclusion}

We implemented a general, end-to-end neural %sequence-to-sequence 
seq2seq model that is able to produce well-formed DRSs with high accuracy (\textbf{RQ1}). Character-level models can outperform word-level models, even though they are not dependent on tokenization and pre-trained word embeddings (\textbf{RQ2}). It is beneficial to rewrite DRS variables to a more general representation (\textbf{RQ3}). Obtaining and employing additional data can benefit performance as well, though it might be better to use an external parser instead of doing a full self-training pipeline (\textbf{RQ4}). 
F-score is only a rough measure for semantic accuracy: Boxer still outperformed our best neural models on a subset of specific semantic phenomena (\textbf{RQ5}). 

We think there are a lot of opportunities for future work. Since the sentences in the PMB data set are relatively short, it makes sense to investigate seq2seq models performing well for longer texts. There are a few promising directions here that could combat the degrading performance on longer sentences. First, the Transformer model \cite{transformer:17} is an interesting candidate for exploration, a state-of-the-art neural model developed for MT that does not have worse performance for longer sentences. Second, a seq2seq model that is able to first predict the general structure of the DRS, after which it can fill in the details, similar to \newcite{neural_drs_gmb:18}, is something that could be explored. A third possibility is a neural parser that tries to build the DRS incrementally, producing clauses for different parts of the sentence individually, and then combining them to a final DRS.

Concerning the evaluation of DRS-parsers, we feel there are a couple of issues that could be addressed in future work. One idea is to facilitate computing F-scores tailored to specific semantic phenomena that are dubbed important, so the evaluation we performed in this paper manually could be carried out automatically.
Another idea is to evaluate the application of DRSs to improve performance on other linguistic or semantic tasks, in which DRSs that capture the full semantics will, presumably, have an advantage. A combination of glass-box and black-box evaluation seems a promising direction here \cite{Bos2008LREC,pmb-LREC:18}.

\section*{Acknowledgements}

This work was funded by the NWO-VICI grant ``Lost in Translation -- Found in Meaning'' (288-89-003). The Tesla K40 GPU used in this work was kindly donated to us by the NVIDIA Corporation. We also want to thank the three anonymous reviewers for their comments.

\bibliography{tacl}
\bibliographystyle{aclnatbib}

\appendix

\end{document}